\documentclass{infor}

\volume{0}
\issue{0}
\pubyear{2024}
\articletype{research-article}
\doi{10.15388/24-INFOR562}

\usepackage[graphicx]{realboxes}
\usepackage[retain-explicit-plus]{siunitx}
\usepackage[export]{adjustbox}
\usepackage{tablefootnote}
\usepackage{lineno,hyperref}
\usepackage{dirtytalk}
\usepackage{color}
\usepackage{mathtools}
\usepackage{comment}
\usepackage{rotating}
\usepackage{cancel}
\usepackage{algorithm}
\usepackage{algpseudocode}
\usepackage{multirow}
\usepackage{multicol}
\usepackage{graphicx}
\usepackage{booktabs}
\usepackage{etoolbox}
\usepackage[multiple]{footmisc}
\usepackage[skip=0.2\baselineskip]{caption}
\usepackage{mathtools}
\usepackage{comment}
\usepackage[T1]{fontenc}
\usepackage{soul,color}
\usepackage{pifont}
\newcommand{\xmark}{\ding{55}}

\usepackage{blindtext}
\usepackage{hyperref}
\usepackage{nameref}

\newcounter{mylabelcounter}

\makeatletter
\newcommand{\labelText}[2]{%
\refstepcounter{mylabelcounter}%
\immediate\write\@auxout{%
 \string\newlabel{#2}{{\unexpanded{#1}}{\thepage}{{\unexpanded{#1}}}{mylabelcounter.\number\value{mylabelcounter}}{}}%
}
}
\makeatother

\makeatletter
\newcommand\footnoteref[1]{\protected@xdef\@thefnmark{\ref{#1}}\@footnotemark}
\makeatother

\begin{document}

%\linenumbers

\begin{frontmatter}

\title{Online detection and infographic explanation of spam reviews with data drift adaptation}
\runtitle{Online detection and infographic explanation of spam reviews}

\author[a]{\inits{F.}\fnms{Francisco} \snm{de Arriba-Pérez}\ead{farriba@gti.uvigo.es}\bio{bio1}}

\author[a]{\inits{S.}\fnms{Silvia} \snm{García-Méndez}\thanksref{c1}\ead{sgarcia@gti.uvigo.es}\bio{bio2}}
\thankstext[type=corresp,id=c1]{Corresponding author.}

\author[b]{\inits{F.}\fnms{Fátima} \snm{Leal}\ead{fatimal@upt.pt}\bio{bio3}}

\author[c,d]{\inits{B.}\fnms{Benedita} \snm{Malheiro}\ead{mbm@isep.ipp.pt}\bio{bio4}}

\author[a]{\inits{J. C.}\fnms{Juan C.} \snm{Burguillo}\ead{J.C.Burguillo@uvigo.es}\bio{bio5}}

\address[a]{Information Technologies Group, \institution{atlanTTic, University of Vigo}, \cny{Spain}}

\address[b]{Research on Economics, Management and Information Technologies, \institution{Universidade Portucalense}, \cny{Portugal}}

\address[c]{ISEP, Polytechnic of Porto, Rua Dr. António Bernardino de Almeida, 4249-015 Porto, Portugal}

\address[d]{\institution{Institute for Systems and Computer Engineering, Technology and Science}, \cny{Portugal}}

\begin{abstract}
Spam reviews are a pervasive problem on online platforms due to its significant impact on reputation. However, research into spam detection in data streams is scarce. Another concern lies in their need for transparency. Consequently, this paper addresses those problems by proposing an online solution for identifying and explaining spam reviews, incorporating data drift adaptation. It integrates (\textit{i}) incremental profiling, (\textit{ii}) data drift detection \& adaptation, and (\textit{iii}) identification of spam reviews employing Machine Learning. The explainable mechanism displays a visual and textual prediction explanation in a dashboard. The best results obtained reached up to \SI{87}{\percent} spam \textit{F}-measure.
\end{abstract}


\end{frontmatter}

\section{Introduction}

Online reviews are a valuable source of information that influences public opinion and directly impacts customers' decision to acquire a product or service \citep{Zhang2018}. However, some reviews are fabricated to promote or undervalue goods and services artificially, \textit{i.e.}, creating spam data \citep{Ana2019,Hutama2022}. Spammers can be humans or bots dedicated to creating deceptive reviews \citep{Silvia2022,Zineb2022}. In this context, spam detection is a critical task in online systems.Spam negatively impacts the user experience and the performance and security of the system \citep{Wang2021}. 

Consequently, a broad set of Machine Learning (\textsc{ml}) methods has been explored for spam detection, mainly supervised learning \citep{crawford2015survey}. In recent years, Natural Language Processing (\textsc{nlp}) techniques \citep{Garcia-Mendez2022} have been adopted to improve the accuracy of spam detection \citep{garg2021}. Given the dynamic nature of the language and behavior of spammers, the challenge is maintaining the effectiveness of spam detection over time, integrating the detection of model drifts in a stream-based environment as data and concept drifts \citep{Wang2019}. While data drifts are related to changes in the input data, concept drifts reflect over time in the predicted target \citep{Duckworth2021}. Specifically, concept drifts in spam detection refer to the changes in the statistical properties of the spam and non-spam entries over time, which can cause the spam detection system to miss-classify reviews. In addition, in a data stream environment, the distribution of input features used to train the spam detection model can change over time, producing data drifts \citep{barddal2017}. Notably, the latter drifts are easier to detect and deal with in a transparent model than in an opaque one \citep{Cano2019}. 

Explainability in spam detection refers to understanding and explaining how a particular text was classified as spam by an automated system \citep{Stites2021}. Therefore, in spam detection, an interpretable mechanism for \textsc{nlp} and concept drift techniques is required to detect spammers in real-time efficiently. According to \cite{crawford2015survey}, scant data stream spam detection research exists. Consequently, this paper contributes to an interpretable online spam detection framework that combines \textsc{nlp} techniques and data drift detectors. The proposed framework achieves high accuracy in spam detection and makes the detection process transparent, allowing users to understand why a review is classified as spam. The evaluation with two experimental data sets presents about \SI{85}{\percent} in the considered evaluation metrics.

The rest of this paper is organized as follows. Section~\ref{sec:related_work} overviews relevant work concerning profiling, classification, data drifts, and explainability for spam detection tasks. Section~\ref{sec:proposed_method} introduces the proposed method, detailing the data processing, stream-based classification procedures, and online explainability. Section~\ref{sec:experimental_results} describes the experimental setup and presents the empirical evaluation results considering the online classification and explanation. Finally, Section~\ref{sec:conclusion} highlights the achievements and future work.

\section{Related work}
\label{sec:related_work}

As previously mentioned, online reviews have become an essential source of information for consumers to make purchasing decisions \citep{Zhang2018,Shaha2022}. However, spam reviews, which are fake or biased reviews, have become a significant problem, leading to distrust and confusion among consumers \citep{Bian2021}. Accordingly, detecting spam reviews is challenging due to the variety of spamming techniques used by spammers; hence, researchers have proposed various approaches for spam review detection \citep{WU2018}. These techniques are based on \textsc{ml} methods \citep{Albayati2019,liu2019method,sun2022} and social network analysis \citep{liu2016,sun2022}. A representative example of the latter is the work by \cite{Rathore2021} on fake reviewer group detection. Their offline graph-based solution, where nodes and edges represent reviewers and products reviewed, respectively, combines the DeepWalk algorithm with semi-supervised clustering. The authors do not perform textual analysis of the reviews, except sentiment analysis.

Spam detection involves large volumes of data, which can be dynamic and continuously changing \citep{Wang2019}. In the case of data streams, not only are reviews continuously arriving, but their statistical properties may change over time, leading to the concept and data drifts \citep{Salih2019}. On the one hand, the volume and speed of online reviews require the adoption of online spam detection techniques \citep{MILLER2014}. On the other hand, outcome explainability is crucial for humans to comprehend, trust, and manage the next generation of cyber defense mechanisms such as spam detection \citep{charmet2022explainable}. Therefore, this related work compares existing works in terms of (\textit{i}) stream-based profile modeling for spam detection, (\textit{ii}) stream-based classification mechanisms, and (\textit{iii}) transparency and credibility in detection tasks.

\subsection{Profiling and classification}

Profiling is the process of modeling stakeholders according to their contributions and interactions \citep{kakar2021value,Silvia2022}. In the case of spam detection, individual profiles are built from the content generated by each stakeholder, humans or bots alike. To overcome information sparsity, the profiles are expected to include side and content information \citep{Faris2019,MOHAWESH2021}, since a richer profile impacts the quality of \textsc{ml} results \citep{Rustam2021}. Mainly, with stream-based modeling, profiles are incrementally updated and refined over time \citep{Veloso2019,Veloso2020}. Concerning online spam detection, the literature considers primary profiling methodologies:

\begin{description}
 \item [Content-based] profiling explores textual features extracted from the text to identify the meaning of the content \citep{SONG2016125,henke2021,MOHAWESH2021}. It can be obtained using linguistic and semantic knowledge or style analysis via \textsc{nlp} approaches. 
 
 \item [User-based] profiling focuses on both the demographic and the behavioral activity of the user \citep{MILLER2014,Nasim2015,liu2016,liu2019method,sun2022}. It contemplates demography information, frequency, timing, and content of posts to distinguish legitimate from spammer users. In addition, exploiting the social graph can be relevant since spammers have many followers or friends who are also suspected of being spammers.
 
\end{description}

Spam detection is a classification task \citep{Vaitkevicius2020,MOHAWESH2021}. The main classification techniques encompass supervised, semi-supervised, unsupervised, and deep learning approaches \citep{crawford2015survey} and can be applied offline or online. While offline or batch processing builds static models from pre-existing data sets, online or stream-based processing computes incremental models from live data streams \citep{Leal2021}. This paper focuses on stream-based environments. Regarding transparency, classification models can be divided into interpretable and opaque. Opaque mechanisms behave as black boxes (\textit{e.g.}, deep learning), and interpretable models are self-explainable (\textit{e.g.}, trees- or neighbor-based algorithms) \citep{Carvalho2019}. Interpretable classifiers explain classification outcomes, clarifying why a given content is false or misleading \citep{Skrlj2021}.

\subsection{Stream-based spam detection approaches}

Social networking has increased spam activity \citep{Kaur2018}. In this context, spam detection approaches have been explored by social networks (\textit{e.g.}, Twitter\footnote{Available at {\scriptsize \url{https://twitter.com}}, May 2024.}, or Facebook\footnote{Available at {\scriptsize \url{https://www.facebook.com}}, May 2024.}) \citep{MILLER2014,Nasim2015,liu2016,sun2022}, email boxes \citep{henke2021}, or crowdsourcing platforms (\textit{e.g.}, Wikipedia\footnote{Available at {\scriptsize \url{https://es.wikipedia.org}}, May 2024.}, Yelp\footnote{Available at {\scriptsize \url{https://yelp.com}}, May 2024.}, and TripAdvisor\footnote{Available at {\scriptsize \url{https://www.tripadvisor.com}}, May 2024.}) \citep{MOHAWESH2021}. Stream mining became the most effective spam detection approach due to the speed and volume of data. It has been explored in the literature using:

\begin{itemize}
 \item \textbf{Data stream clustering} approaches. \cite{MILLER2014} treated spam detection as an anomaly prediction problem. The proposed solution identifies spammers on Twitter using account information and streaming tweets employing stream-based clustering algorithms. \cite{Nasim2015} followed the same methodology, creating clusters of tweets and considering outliers as spam. \cite{SONG2016125} proposed a new ensemble approach named Dynamic Clustering Forest (\textsc{dcf}) for the classification of textual streams, which combines decision trees and clustering algorithms.

 \item \textbf{Data stream classification} for spam detection. \cite{sun2022} proposed a near real-time Twitter spam detection system employing multiple classification algorithms and parallel computing. 

 \item \textbf{Outlier detection for stream data}. \cite{liu2019method} proposed solution identifies outlier reviews, analyzes the differences between the patterns of product reviews, and employs an isolation forest algorithm.
\end{itemize}

\subsubsection{Drifts in spam detection}

Model drift occurs when the performance of an \textsc{ml} model loses accuracy over time \citep{Ma2023}. The literature identifies two types of drifts: (\textit{i}) data drifts and (\textit{ii}) concept drifts. While data drift occurs when the characteristics of the incoming data change, in concept drifts, both input and output distributions present modifications over time \citep{Desale2023}. According to \cite{Gama2014}, concept drift detection methods can be divided into three categories: (\textit{i}) sequential analysis, (\textit{ii}) statistical analysis, and (\textit{iii}) sliding windows. In addition, for \cite{lu2018learning}, drift detection involves four stages: (\textit{i}) data retrieval, (\textit{ii}) data modeling, (\textit{iii}) test statistics calculation, and (\textit{iv}) hypothesis test. 

\cite{liu2016} proposed and applied two online drift detection techniques to improve the classification of Twitter spam reviews: (\textit{i}) fuzzy-based redistribution and (\textit{ii}) asymmetric sampling. While the fuzzy-based redistribution technique explores information decomposition, asymmetric sampling balances the size of classes in the training data. \cite{SONG2016125} analyzed the distribution of textual information to identify concept drifts in a textual data classification approach. Moreover, \cite{MOHAWESH2021} employed a comprehensive analysis to address concept drift in detecting fake Yelp reviews. Finally, \cite{henke2021} monitored feature evolution based on the similarity between feature vectors to concept drifts in emails. The solution performs spam classification and concept drift detection as parallel and independent tasks.

In contrast to the previous drift detection works, the current approach adopts self-explainable models to provide explanations, increasing classification quality and user trust.

\subsubsection{Explainability}

Explainable spam detection refers to explaining why an input was classified as spam. It promotes transparency and clarity, detailing why a particular review was flagged as spam \citep{Stites2021}. Accordingly, interpretable models, such as rule-based systems or decision trees, can explain their reasoning, enhancing trust, reducing bias, and helping to discover additional insights \citep{Rudin2019}. In addition, \textsc{nlp} enriches the explanations by adding a textual description \citep{Upadhyay2021}. Explainable spam detection has been explored in the literature using Local Interpretable Model Agnostic Explanation (\textsc{lime}) \citep{Ribeiro:2016} and Shapley Additive Explanations (\textsc{shap}) \citep{reis2019explainable,han2022explainable,zhang2022explainable}.

The literature shows that existing explainable detectors of fake content in online platforms adopt essentially supervised classification and implement offline processing \citep{crawford2015survey,henke2021}. Therefore, this paper intends to address this problem by proposing an online solution for identifying and explaining spam reviews, incorporating data drift detection and adaptation.

\subsection{Research contribution}
\label{sec:research_contribution}

The literature review shows a research gap in detecting data drifts and explaining the classification of textual reviews as spam in real time. In this respect, \cite{RAO2021} identifies spam drift detection as a challenge requiring more research. Table \ref{tab:comparison} provides an overview of the above works considering the data domain, profiling (user- and content-based), spam detection, drift detection, and explainability. 

Therefore, this work contributes with an online explainable classification method to recognize spam reviews and, thus, promote trust in digital media. The solution employs data stream processing, updating profiles, and classifying each incoming event. First, user profiles are built using user- and content-based features engineered through \textsc{nlp}. Then, the proposed system monitors the incoming streams to detect data drifts using static and sliding windows. Tree-based classifiers are exploited to obtain an interpretable stream-based classification for classification. Finally, the proposed method provides the user with a dashboard combining visual data and natural language knowledge to explain why an incoming review was classified as spam.

\begin{table}[!htbp]
\centering
\footnotesize
\caption{Comparison of stream-based spam and drift detection approaches\tablefootnote{DT - Decision Tree, LR - Logistic Regression, PNN - Perceptron Neural Network, RF - Random Forest, SVM - Support Vector Machine.}.}
\label{tab:comparison}
\begin{tabular}{lccccc}
\toprule
\textbf{Authorship} & \textbf{Domain} &\textbf{Profiling} & \textbf{Spam} & \textbf{Drift} & \textbf{Explainability} \\
 & & & \textbf{Detection} & \textbf{Detection} & \\
\midrule
\multirow{2}{*}{\cite{liu2016}} & \multirow{2}{*}{Twitter} & Content & Classification& \multirow{2}{*}{Data} & \multirow{2}{*}{\xmark} \\
 & & User & (Multiple)\\
\multirow{2}{*}{\cite{SONG2016125}} & \multirow{2}{*}{Spam} & \multirow{2}{*}{Content} & Clustering & \multirow{2}{*}{Concept} & \multirow{2}{*}{\xmark} \\
 & & & (\textsc{dt}) \\
\multirow{2}{*}{\cite{MOHAWESH2021}} & \multirow{2}{*}{Yelp} & \multirow{2}{*}{Content} & Classification & \multirow{2}{*}{Concept} & \multirow{2}{*}{\xmark} \\
 & & & (\textsc{lr}, \textsc{pnn}, \textsc{svm}) \\
\multirow{2}{*}{\cite{henke2021}} & \multirow{2}{*}{Email} & \multirow{2}{*}{Content} & Classification & \multirow{2}{*}{Concept} & \multirow{2}{*}{\xmark} \\
 & & & (\textsc{svm}) \\
\midrule
\multirow{2}{*}{\bf Proposed solution} & \multirow{2}{*}{Yelp} & Content & Classification & \multirow{2}{*}{Data} & \multirow{2}{*}{\checkmark} \\
& & User & (\textsc{dt}, \textsc{rf})\\
\bottomrule
\end{tabular}
\end{table}

As previously explained, concept drift refers to changes in the predicted target over time (\textit{i.e.}, changes in the statistical properties of the spam and non-spam entries), while data drift focuses on input data variations (\textit{i.e.}, changes in the input features used to train the spam detection model). This work focuses on data drift detection, considering its relationship with the transparency of the model. Specifically, detecting data drifts and associated characteristics helps provide richer information to end users via the explainability dashboard. Although no other work has explored the Yelp dataset for data drift and spam detection, work on other topics, such as sentiment analysis, indicates its suitability
 \citep{Chumakov2023,Madaan2023,Wu2023}.

\section{Method}
\label{sec:proposed_method}

\begin{figure}[!htb]
\centering
\includegraphics[scale=0.13]{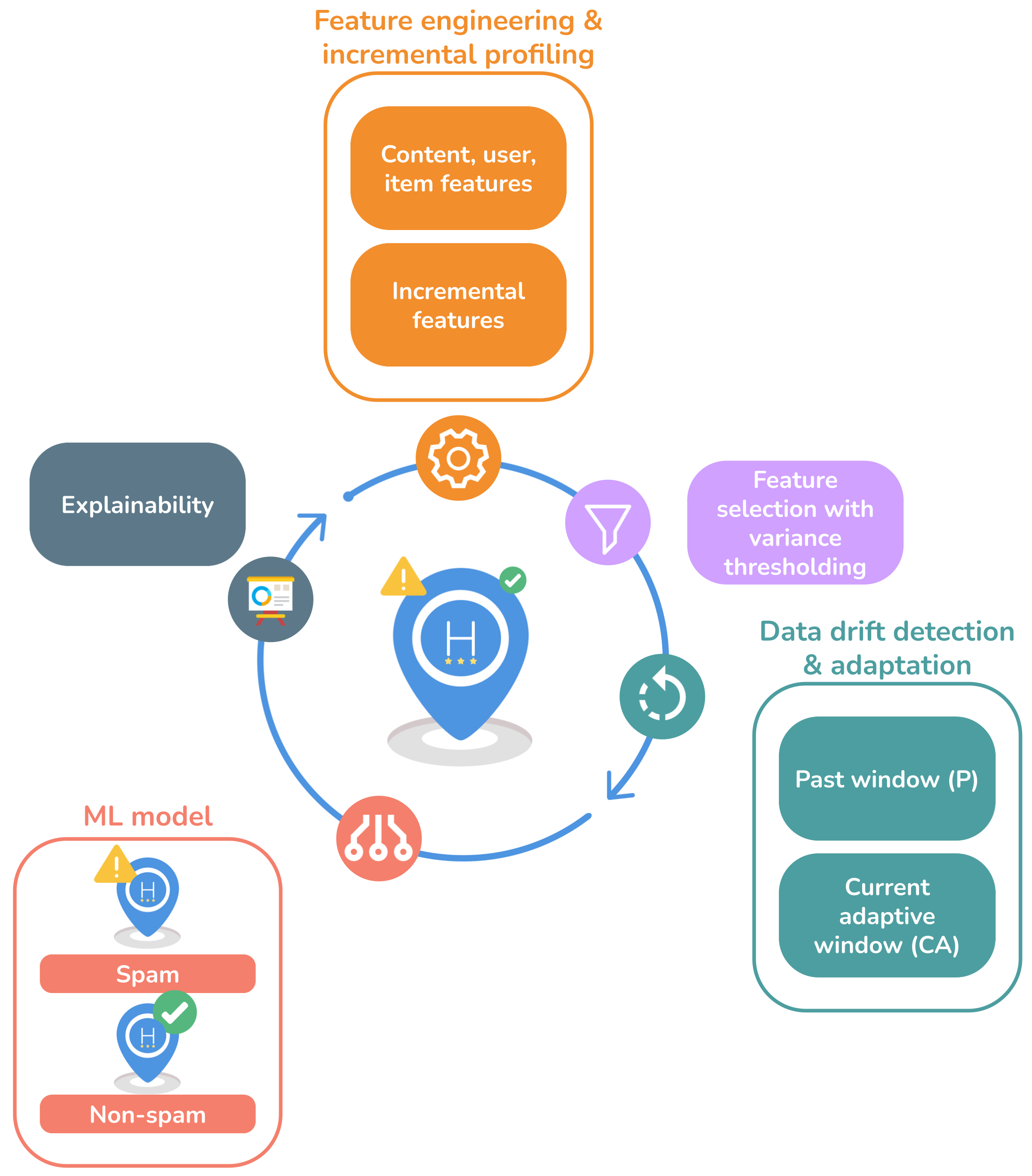}
\caption{\label{fig:scheme}Data stream classification pipeline.}
\end{figure}

The proposed method explores online reviews for stream-based spam classification with drift detection. In addition, it explores self-explainable \textsc{ml} models for transparency. Hence, the data stream classification pipeline, represented in Figure \ref{fig:scheme}, comprises: (\textit{i}) feature engineering \& incremental profiling (Section \ref{sec:feature_engineering}), (\textit{ii}) feature selection (Section \ref{sec:feature_selection}), (\textit{iii}) data drift detection \& adaptation (Section \ref{sec:data_drift}), (\textit{iv}) \textsc{ml} classification (Section \ref{sec:ml_classification}), and (\textit{v}) explainability (Section \ref{sec:explainability}).

\subsection{Feature engineering \& incremental profiling}
\label{sec:feature_engineering}

The proposed solution processes the content of the reviews with the help of \textsc{nlp} techniques. The content-based features extracted represent relevant linguistic (morphological, syntactical, and semantic) attributes of the reviews. The engineered features are the ratio of adjectives, adverbs, interjections, nouns, pronouns, punctuation marks, verbs, char, word, difficult word, and \textsc{url} counters. Moreover, the system also considers the emotional charge of the content (\textit{i.e.}, anger, fear, happiness, sadness, and surprise). The same applies to the polarity charge among negative, neutral, and positive sentiments. More sophisticated linguistic features include readability, using the Flesch readability score, the McAlpine \textsc{eflaw} score\footnote{A value higher than 25 points is unfavorable.}, and the reading time. In the end, the content itself, \textit{i.e.}, the words, are analyzed through word-grams. The char-grams were discarded due to their low scalability in online operation. These content-based features are then used to incrementally build the corresponding user values to update the user profiles. Additionally, incremental relational item features are computed by building a graph of item and user nodes connected by edges containing the corresponding incremental engineered features of the user-reviewed items.

\subsection{Feature selection}
\label{sec:feature_selection}

Feature selection reduces the feature space dimension by choosing the most relevant features for the classification and contributes to improving the quality of the input data. The adopted selection technique relies on feature variance to discard those with variance lower than a configurable threshold, as suggested by the literature \citep{Engelbrecht2019,Treistman2022}. In the case of online classification, where the arriving data may evolve with time, the selection of representative features must be performed continuously or periodically.

\subsection{Data drift detection and adaptation}
\label{sec:data_drift}

The variability of real data over time may affect the performance of \textsc{ml} models, namely the values of evaluation metrics (\textit{e.g.}, accuracy, precision, recovery, etc.). However, the source of the problem may be due to data drifts, concept drifts, ineffective hyper-parameter optimization, and/or class imbalance. 

Thus, the proposed system continuously monitors the incoming stream for data drifts and, periodically, under-samples and optimizes the hyperparameters, using two windows: the past (\textsc{p}) static window and the current adaptive (\textsc{ca}) sliding window, holding \textit{n} and \textit{w} samples, respectively. 

\begin{figure*}[!htbp]
\centering
\includegraphics[scale=0.14]{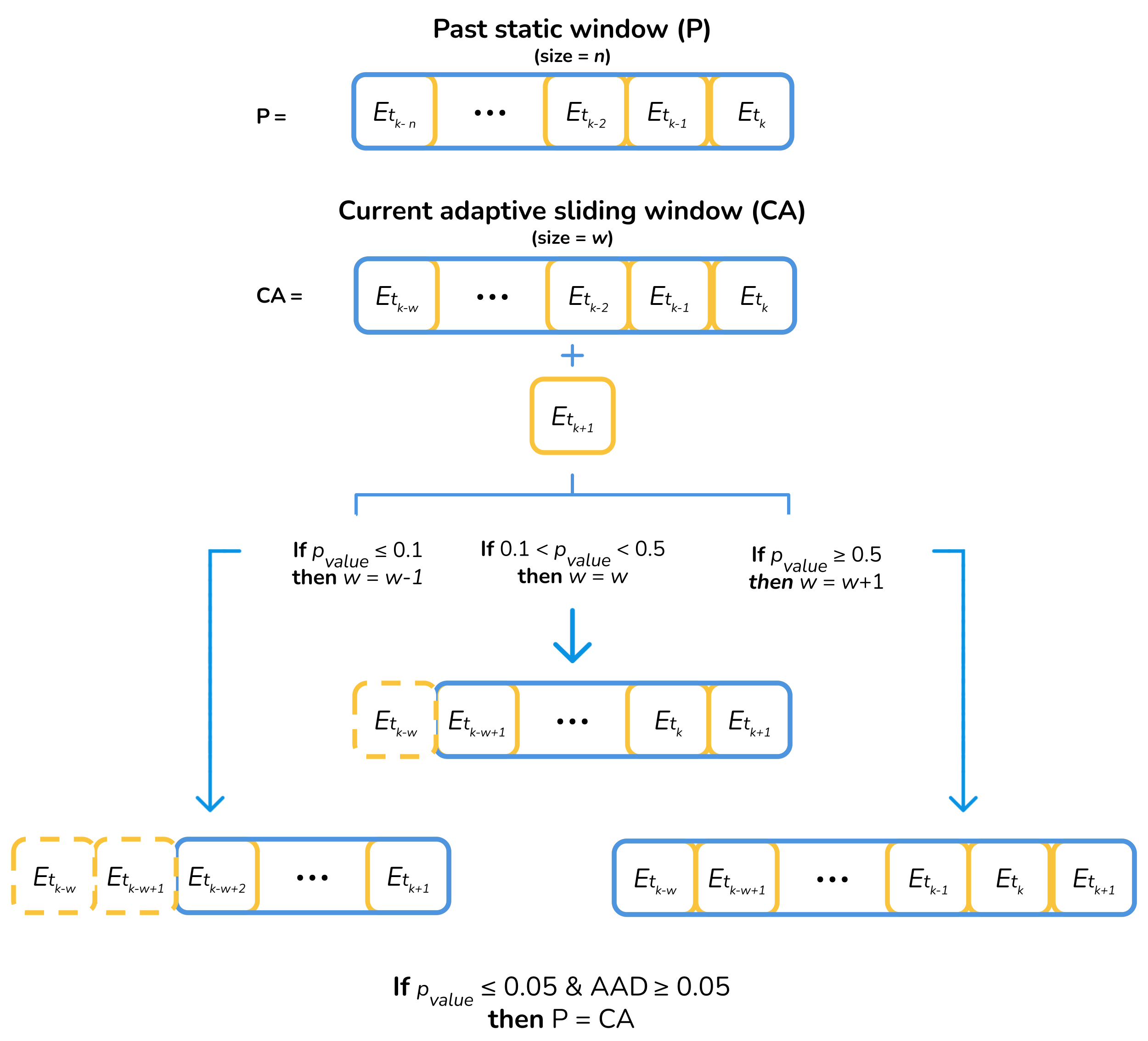}
\caption{\label{fig:data_drift}Data drift detection and adaptation.}
\end{figure*}

The data drift detector starts operating when the cold start ends, and the \textsc{p} window is initialized with the expected \textit{n} samples. The detector identifies a data drift whenever: (\textit{i}) the inter-window word-gram \textit{p-value} is lower\footnote{In a modern language, the most frequent words in a text are not expected to vary over time, leading to \textit{p-values} greater than \num{0.05}. However, the contents and the words within spam texts are anticipated to vary over time, resulting in \textit{p-values} below \num{0.05}.} than \num{0.05}, and (\textit{ii}) the inter-window \textit{absolute accuracy difference} (\textsc{aad}) is higher than \num{0.05}. Algorithm \ref{alg:data_drift} details the data drift detection and adaptation process. The threshold values of \num{0.05}, \num{0.1}, and \num{0.5} were inspired by the works by \citeauthor{Solari2017,Giovanni2020,Ritu2021}, respectively. Figure \ref{fig:data_drift} illustrates this process. The data drift detector works as follows:

\begin{itemize}

 \item Calculates the word-gram frequency matrices (\textit{i.e.}, the columns represent the word-grams and the rows, the entries) for the \textsc{p} and \textsc{ca} windows.
 
 \item Sum\_wordgrams method transforms the latter matrices into vector format (a vector for \textsc{p} and a vector for \textsc{ca}) by summing the word-gram frequency for all entries.

 \item Discards the columns with a frequency lower than 6 in both sum\_wordgrams vectors.

 \item Computes the \textit{p-value} between the word-grams frequency vectors of \textsc{p} and \textsc{ca} windows.
 
 \item Computes the inter-window \textsc{aad}.

  \item Updates the size of the \textsc{ca}:

 \begin{itemize}

     \item If the \textit{p-value} $\leq$ \num{0.1}, the \textsc{ca} windows size decrements by one.
    
     \item If the \textit{p-value} > \num{0.1} and \textit{p-value} < \num{0.5}, the \textsc{ca} windows size remains unchanged.
     
     \item If the \textit{p-value} $\geq$ \num{0.5}, the \textsc{ca} windows size increments by one.

 \end{itemize}

 \item Identifies a data drift when the inter-window word-gram \textit{p-value} is lower (or equal) and the inter-window \textsc{aad} is higher (or equal) than \num{0.05}. Then, it replaces the \textsc{p} with the \textsc{ca} window and recalculates the optimal hyperparameters. The hyperparameter\_computation method applies an exhaustive search technique over the configuration parameters listed in Figure \ref{fig:listings}. Ultimately, the \textsc{ml} model is trained using the ml\_update function with the hyperparameters selected and the \textsc{ca} samples.
 
\end{itemize}

\begin{algorithm*}[!htbp]
 \captionsetup{font=small} 
 \small
 \caption{\label{alg:data_drift}: {\bf Data drift detection and classification}}
 \begin{algorithmic}[0]
 \Function{main}{n,ml\_model\_name} \hfill{\%n is the configurable cold start threshold, and the name of the model used (see the list provided in Section \ref{sec:ml_classification})}
 \State $P = []$ \hfill{\%Past static window}
 \State $CA = []$ \hfill{\%Current adaptive sliding window}
 \State $list\_actual = []$ \hfill{\%List with actual values}
 \State $list\_predicted = []$ \hfill{\%List with predicted values}
 
 \State $k=0$ \hfill{\%Sample counter}

 \State $system.listener(sample,drift\_analysis)$ \hfill{\%The system waits for the arrival of a new sample to call the drift analysis function}
\EndFunction

\Function{drift\_analysis}{sample}

 \If{$k < n$} \hfill{\%Warm operation}
 \State $P.append(sample.wordgrams)$
 \EndIf

 \State $CA.append(sample.wordgrams)$
 \State $p_{value} = 0$
 \State $AAD = 0$
 \State $k=k+1$
 \If{$k == n$}
 \State $acc_p = accuracy(list\_actual,list\_predicted)$
 \EndIf
 \If{$k \geq n$}
 \State $ca\_vector = CA.sum\_wordgrams()$
 \State $p\_vector = P.sum\_wordgrams()$
 \State $p_{value} = chi2(ca\_vector, p\_vector)$
  \State $acc_{ca} = accuracy(list\_actual[-len(CA)\colon],list\_predicted[-len(CA)\colon])$
 \State $AAD = abs(acc_p - acc_{ca})$
 \If{$p_{value} \leq 0.1$}
 \State $CA=CA[2\colon]$
 \EndIf
 \If{$p_{value} > 0.1$ {\bf and} $p_{value} <	0.5$}
 \State $CA=CA[1\colon]$
 \EndIf

 \If{$p_{value} \leq 0.05$ {\bf and} $AAD \geq 0.05$}
 \State $P = CA$
 \State $parameters_{updated} = hyperparameter\_computation(ml\_model\_name,CA)$
 \State $ml\_model=ml\_update(parameters_{updated})$
 \State $acc_p = accuracy(list\_actual[-len(P)\colon],list\_predicted[-len(P)\colon]$

 \EndIf

 \EndIf
 
 \State $predicted, actual = ml\_classification\_step(ml\_model,sample)$
 \State $list\_actual.append(actual)$
 \State $list\_predicted.append(predicted)$

 \State $sample= input\_new\_sample()$
  
 \EndFunction
 \end{algorithmic}
\end{algorithm*}

\subsection{ML classification}
\label{sec:ml_classification}

The following online \textsc{ml} algorithms were used as they exhibited good performance in similar classification problems \citep{liu2016,SONG2016125,sun2022}.

\begin{itemize}
 
 \item \textbf{Hoeffding Tree Classifier} (\textsc{htc}) \citep{Pham2017} is the basic decision tree model for online learning.

 \item \textbf{Hoeffding Adaptive Tree Classifier} (\textsc{hatc}) \citep{Stirling2018} monitors branches and replaces them based on their performance.
 
 \item \textbf{Adaptive Random Forest Classifier} (\textsc{arfc}) \citep{Gomes2017} is an ensemble of trees with diversity induction through random re-sampling and concept drift detection. The prediction results are obtained using majority voting.
 
\end{itemize}

The algorithmic performance assessment follows the prequential evaluation protocol \citep{Gama2013} and considers accuracy, macro- and micro-averaging \textit{F}-measure, and run-time metrics.

\subsection{Explainability}
\label{sec:explainability}

\begin{figure}[!htb]
\centering
\includegraphics[width=0.80\textwidth]{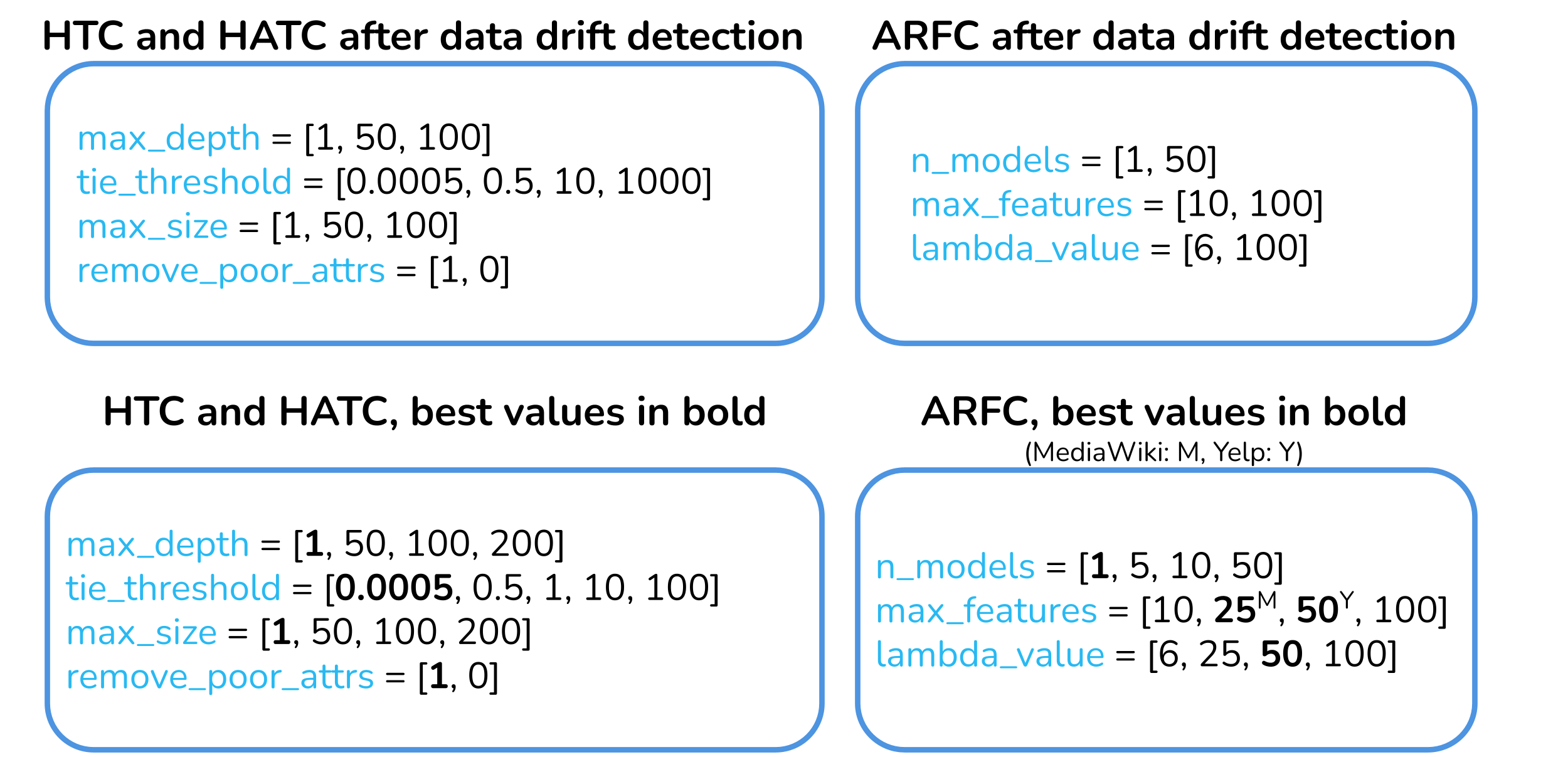}
\caption{\label{fig:listings}Model hyperparameter configuration (best values in bold).}
\end{figure}

This module provides information about the most relevant features for the classification, \textit{i.e.}, those with a frequency of appearance greater than a configurable threshold. This information is extracted from the estimators of the tree models used (see Figure \ref{fig:listings}): \textsc{htc}  single-model estimator \citep{Pham2017}, \textsc{hatc} single-model estimator \citep{Stirling2018} and \textsc{arfc} multi-model estimator \citep{Gomes2017}. The predictions regarding the most relevant features and data drift detection are described in natural language. Furthermore, the decision tree path followed is also provided, along with an automatic description obtained from a Large Language Model.

\section{Experimental results}
\label{sec:experimental_results}

This section describes the experimental data set (Section \ref{sec:experimental_dataset}) and the implementation of the different modules\footnote{Code available at \url{https://github.com/nlpgti/data_drift}.}: (\textit{i}) feature engineering \& incremental profiling (Section \ref{sec:feature_engineering_results}), (\textit{ii}) feature selection (Section \ref{sec:feature_selection_results}), and (\textit{iii}) data drift detection \& adaptation (Section \ref{sec:data_drift_results}). The classification and explainability results are detailed in Section \ref{sec:ml_classification_results} and Section \ref{sec:explainability_results}, respectively.

The experiments contemplate four stream classification scenarios, incorporating feature selection, hyperparameter optimization\footnote{Hyper-parameter optimization was performed with the \SI{0.005}{\percent} of the experimental samples in Section \ref{sec:ml_classification_results}.} and incremental accuracy updating.

\begin{description}
 
\item \textbf{Scenario 1}. The data stream classification runs on a single processing thread.

\item \textbf{Scenario 2}. The data stream classification runs on a range of \num{10}-\num{20} parallel threads based on the workload to reduce the experimental run-time. To preserve the original data distribution, the chronologically ordered data stream was divided into consecutive sub-streams, and then, each sub-stream was processed in a dedicated thread.

\item \textbf{Scenario 3}. The data stream classification includes data drift detection \& adaptation and runs according to scenario 2.

\item \textbf{Scenario 4}. The data stream classification runs on a single processing thread with data drift detection \& adaptation\footnote{Due to time limitations, this scenario will only be applied with the best classifier so far.}.

\end{description}

All experiments were performed using a server with the following hardware specifications:
\begin{itemize}
 \item \textbf{Operating System}: Ubuntu 18.04.2 LTS 64 bits
 \item \textbf{Processor}: Intel\@Core i9-10900K \SI{2.80}{\giga\hertz}
 \item \textbf{RAM}: \SI{96}{\giga\byte} DDR4 
 \item \textbf{Disk}: \SI{480}{\giga\byte} NVME + \SI{500}{\giga\byte} SSD
\end{itemize}

\subsection{Experimental data set}
\label{sec:experimental_dataset}

The Yelp data set\footnote{Available at {\scriptsize \url{https://www.kaggle.com/datasets/abidmeeraj/yelp-labelled-dataset?select=Labelled+Yelp+Dataset.csv}}, May 2024.} is composed of \num{359052} leisure activity entries between October 2004 and January 2015, distributed between \num{36885} and \num{322167} samples of spam and non-spam content, respectively (see Table \ref{tab:dataset_distribution}). Moreover, the MediaWiki data set\footnote{Available from the corresponding author on reasonable request.} contains contributions to travel wikis between August 2003 and June 2020. It is composed of \num{319856} entries, distributed between \num{24877} and \num{249979} samples of spam and non-spam content, respectively (see Table \ref{tab:dataset_distribution}).

\begin{table}[!htbp]
\centering
\caption{\label{tab:dataset_distribution}Distribution of classes in the experimental data sets.}
\begin{tabular}{llS[table-format=6.0]}
\toprule 
\bf Data set & \textbf{Class} & \multicolumn{1}{c}{\textbf{Number of entries}}\\ \midrule
\multirow{3}{*}{Yelp} & Spam & 36885\\
& Non-spam & 322167\\
\cmidrule{2-3}
& \bf Total & 359052\\
\midrule
\multirow{3}{*}{MediaWiki} & Spam & 24877\\
& Non-spam & 294979\\
\cmidrule{2-3}
& \bf Total & 319856\\
\bottomrule
\end{tabular}
\end{table}

\subsection{Feature engineering \& incremental profiling}
\label{sec:feature_engineering_results}

This section details the implementations and \textsc{nlp} techniques used to create the classification features. Table \ref{tab:features}, Table \ref{tab:features_incremental}, and Table \ref{tab:features_graph} detail the content features, the incremental user features, and the incremental item features for Yelp and MediaWiki data sets, respectively.

Most ratio and counter features in Table \ref{tab:features} (features 1, 2, 7, 9, 11, 12, 15 in Table \ref{tab:features}) are computed using the spaCy\footnote{Available at {\scriptsize \url{https://spacy.io}}, May 2024.} tool to gather their grammatical category (\texttt{\small token.pos\_} feature). The char and word count (features 3 and 16, respectively) have been directly calculated with the Python \texttt{\small len} function\footnote{For feature 16, the text was first separated into word tokens.}. The \textsc{url} count (feature 14) was computed using a regular expression\footnote{Available at {\scriptsize \url{https://bit.ly/3N4GNM3}}, May 2024.}. The emotion (feature 5) and polarity (feature 10) are calculated using \texttt{\small Text2emotion}\footnote{Values between 0 and 1. Available at {\scriptsize \url{https://pypi.org/project/text2emotion}}, May 2024.} and \texttt{\small TextBlob}\footnote{Values between -1 and 1. Available at {\scriptsize \url{https://pypi.org/project/spacytextblob}}, May 2024.}, respectively. The rating-polarity deviation is computed as the difference between those values after moving the polarity to a Likert scale\footnote{Polarity\_likert = 2.5*(polarity + 1).} (feature 18). The system uses \texttt{\small Textstat}\footnote{Available at {\scriptsize \url{https://pypi.org/project/textstat}}, May 2024.} for the readability (features 4, 6 and 8) and reading time (feature 13). Word-grams (single and bi-words, feature 17) are obtained with \texttt{\small CountVectorizer}\footnote{Available at {\scriptsize \url{https://scikit-learn.org/stable/modules/generated/sklearn.feature_extraction.text.CountVectorizer.html}}, May 2024.} with the \textsc{hatc} model as the meta-transformer, and using the following parameters: \texttt{\small max\_df=0.7}, \texttt{\small \small min\_df=0.1}\footnote{For the MediaWiki data set, min\_df=0.01 since the reviews are shorter.}. For the word-grams generation, the review is pre-processed, removing non-textual characters (numbers, punctuation marks, and subsequent blank spaces), stop words\footnote{Available at {\scriptsize \url{https://gist.github.com/sebleier/554280}}, May 2024.}, and \textsc{url} instances. Then, the review text is lemmatized with spaCy using the \texttt{\small en\_core\_web\_md} model\footnote{Available at {\scriptsize \url{https://spacy.io/models/en}}, May 2024.}. The drift detector exclusively uses the inter-window word-grams \textit{p-value} variations.

Table \ref{tab:features_incremental} and Table \ref{tab:features_graph} summarize the user incremental features (58 features) and item incremental features (92 features) generated from the content-based features in Table \ref{tab:features}. The user engineered features of Table \ref{tab:features_incremental} and Table \ref{tab:features_graph} correspond to the incremental average $f_{avg_{t_{k}}}$ given by Equation \ref{eq:incremental_average} and the incremental maximum $f_{max_{t_{k}}}$ given by Equation \ref{eq:incremental_maximun}, where $f$ represents the feature and $[f_{t_o}, f_{t_1}, ..., f_{t_k}]$ the past feature data per user.

\begin{equation}
\label{eq:incremental_average}
f_{avg_{t_{k}}} = \frac{\sum_{i=0}^k{f_{t_{i}}}}{k}
\end{equation}

\begin{equation}
\label{eq:incremental_maximun}
f_{max_{t_{k}}} = \max_{i}{f_{t_{i}}}
\end{equation}

\begin{table}[!htbp]
\centering
\footnotesize
\caption{\label{tab:features} Content-based features explored per experimental data set.}
\begin{tabular}{ccp{3.2cm}p{5cm}c} 
\toprule
\bf Data set & \bf ID & \bf Name & \bf Description & \bf Type \\\midrule
\multirow{19}{*}{\rotatebox[origin=c]{90}{Common}} & 1 & Adjective ratio & Ratio of adjectives in the content & \multirow{19}{*}{\rotatebox[origin=c]{90}{Engineerd (Eng.)}}\\
& 2 & Adverb ratio & Ratio of adverbs in the content\\
& 3 & Char count & Number of characters in the content\\
& 4 & Difficult word count & Number of the difficult words in the content\\
& 5 & Emotion (anger, fear, happiness, sadness, surprise) & Load of the different emotions in the content\\
& 6 & Flesch readability & Readability score of the content\\
& 7 & Interjection ratio & Ratio of interjections in the content\\
& 8 & McAlpine \textsc{eflaw} readability & Readability score of the content for non-native English speakers\\
& 9 & Noun ratio & Ratio of nouns in the content\\
& 10 & Polarity & Sentiment of the content\\
& 11 & Pronoun ratio & Ratio of pronouns in the content\\
& 12 & Punctuation ratio & Ratio of punctuation marks in the content\\
& 13 & Reading time & Content reading time\\
& 14 & \textsc{url} count & Number of \textsc{url} in the content\\
& 15 & Verb ratio & Ratio of verbs in the content\\
& 16 & Word count & Number of words in the content\\
& 17 & Word $n$-grams & Single and bi-words grams\\
\midrule
\multirow{4}{*}{\rotatebox[origin=c]{90}{Yelp}} & 18 & Rating-polarity deviation & Rating deviation concerning the polarity of the content & Eng.\\ \cmidrule{5-5}
& 19 & Review rating & Rating of the review & Raw\\
\midrule
\multirow{11}{*}{\rotatebox[origin=c]{90}{MediWiki}} & 20 & Bot flag & The user is a bot & \multirow{11}{*}{\rotatebox[origin=c]{90}{Raw}}\\
& 21 & Deleted flag & Part of the revision content is hidden\\
& 22 & New flag & It is the first revision of a page\\ 
& 23 & Revert flag & The revision was reverted & \\
& 24 & Size difference & Difference in the number of characters added and deleted in the revision\\
& 25 & Edit quality & False/true damaging \& good faith probability\\ 
& 26 & Item quality & \textsc{a}, \textsc{b}, \textsc{c}, \textsc{d}, \textsc{e} probability\\
& 27 & Article quality & \textsc{ok}, attack, vandalism, \textsc{wp10b}, \textsc{wp10c}, \textsc{wp10fa}, \textsc{wp10ga}, \textsc{wp10start}, \textsc{wp10stub} probability\\
\bottomrule
\end{tabular}
\end{table}

\begin{table}[!htbp]
\centering
\footnotesize
\caption{\label{tab:features_incremental} User engineered features for both experimental data sets.}
\begin{tabular}{cp{3cm}p{5.5cm}} 
\toprule
\bf ID & \bf Name & \bf Description \\\midrule
\{28, 81\} & User features & Incremental average and maximum per user regarding features \numrange{1}{27} in Table \ref{tab:features}.\\
82 & User post count & Cumulative number of posts per user.\\
83 & User spam tendency & Known spamming behavior per user.\\
84 & User posting antiquity & Posting antiquity per user (in weeks).\\
85 & User posting frequency & Weekly posting frequency per user.\\
\bottomrule
\end{tabular}
\end{table}

\begin{table}[!htbp]
\centering
\footnotesize
\caption{\label{tab:features_graph} Item engineered features for both experimental data sets.}
\begin{tabular}{cp{3cm}p{5.5cm}} 
\toprule
\bf ID & \bf Name & \bf Description \\\midrule
\{86, 139\} & Item features & Incremental average and maximum per item regarding features \numrange{1}{27} in Table \ref{tab:features}.\\
\{140, 177\} & Item and rating features & Incremental average and maximum per item and rating regarding features \numrange{1}{19} in Table \ref{tab:features}.\\
\bottomrule
\end{tabular}
\end{table}

\subsection{Feature selection}
\label{sec:feature_selection_results}

To reduce the feature space dimension, the variance of the features in Table \ref{tab:features} and Table \ref{tab:features_incremental} is analyzed with the help of the \texttt{\small VarianceThreshold}\footnote{Available at {\scriptsize \url{https://riverml.xyz/0.11.1/api/feature-selection/VarianceThreshold}}, May 2024.} from River 0.11.1\footnote{Available at {\scriptsize \url{https://riverml.xyz/0.11.1}}, May 2024.}. The threshold is set to \num{0}, the default value. In the case of Yelp, only feature 14 in Table \ref{tab:features} and its incremental versions in Table \ref{tab:features_incremental} and Table \ref{tab:features_graph} were discarded. The discarded MediaWiki features include features 21 and 22 in Table \ref{tab:features} and their incremental versions in Table \ref{tab:features_incremental} and Table \ref{tab:features_graph}, along with the incremental version of feature 20 in Table \ref{tab:features_incremental}. All remaining features passed the threshold and were, thus, considered relevant for the classification.

\subsection{Data drift detection and adaptation}
\label{sec:data_drift_results}

While standard online \textsc{ml} models can adapt to data changes over time, they are still affected by data drift, also known as covariate shift. To address this issue, scenario 3 incorporates data drift detection \& adaptation. Moreover, it defines that: (\textit{i}) the cold start spans over the first \num{500} samples, corresponding to the initial width of the \textsc{p} window; (\textit{ii}) the maximum width of \textsc{ca} sliding windows is \num{2000} samples. The proposed data drift detector determines the inter-window word-gram \textit{p-value} and the inter-window \textsc{aad}, using the \texttt{\small Chi2ContingencyResult} function\footnote{Available at {\scriptsize \url{https://docs.scipy.org/doc/scipy/reference/generated/scipy.stats.chi2_contingency.html}}, May 2024.} and the \texttt{\small accuracy\_score} function\footnote{Available at {\scriptsize \url{https://scikit-learn.org/stable/modules/generated/sklearn.metrics.accuracy_score.html}}, May 2024.}, respectively.

\begin{figure*}[!htbp]
\centering
\includegraphics[width=0.9\textwidth]{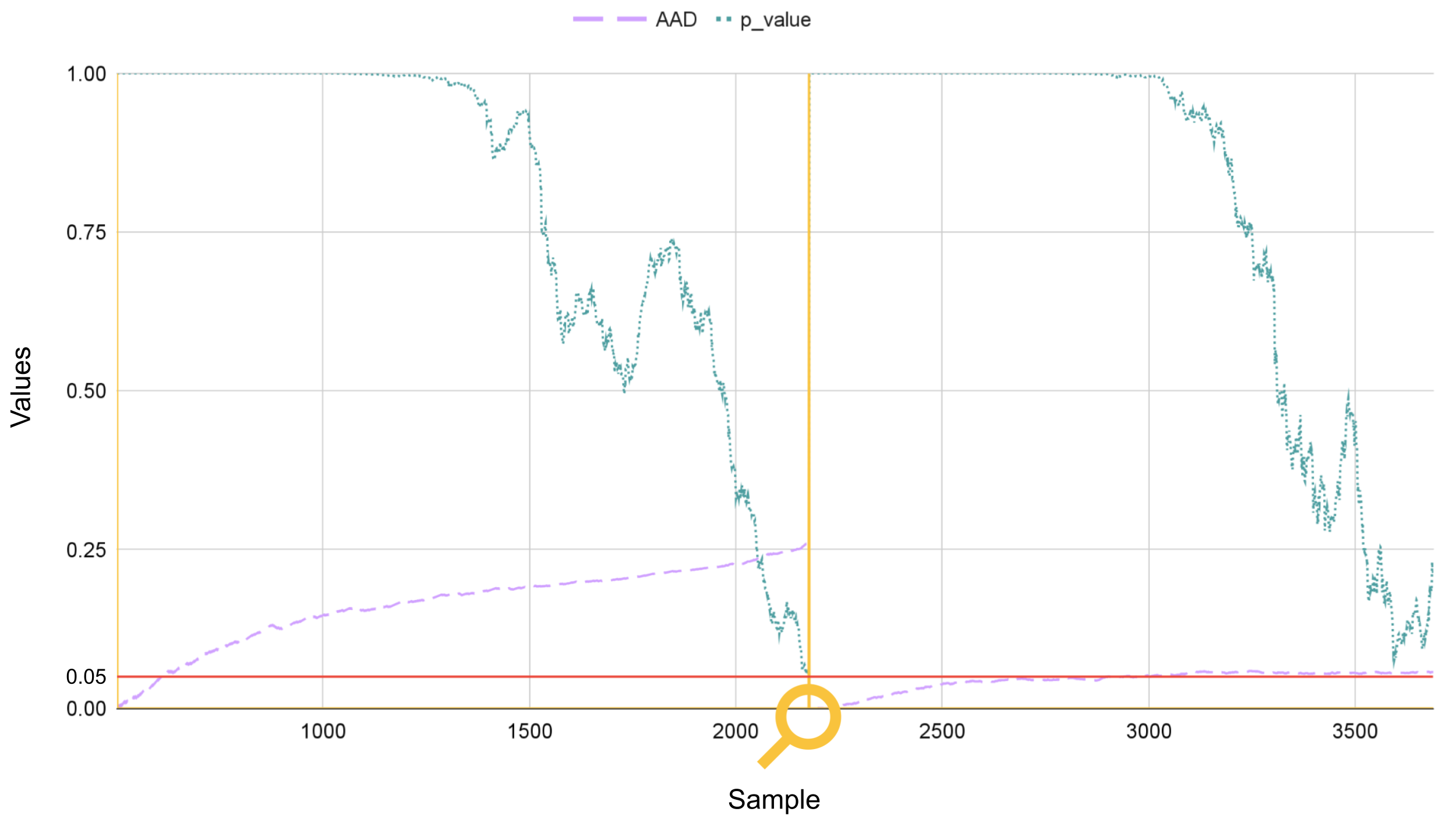}
\caption{\label{fig:accuracy_variation}Data drift detection \& adaptation based on inter-window \textsc{aad} and word-gram \textit{p-value}.}
\end{figure*}

Figure \ref{fig:accuracy_variation} shows the evolution of the inter-window \textsc{aad} and word-gram \textit{p-value}. The lens marks the detected data drift when \textit{p-value} drops to \num{0.05}, and \textsc{aad} is above \num{0.05}.

Once a drift is identified, the hyperparameter optimization starts. 
This process, which is the most time demanding, employs \texttt{\small GridSearch}\footnote{Available at {\scriptsize \url{https://scikit-learn.org/stable/modules/generated/sklearn.model_selection.GridSearchCV.html}}, May 2024.} with reduced configuration parameters (see Figure \ref{fig:listings}).

\subsection{ML classification}
\label{sec:ml_classification_results}

The selected classification techniques include 
 \textsc{htc}\footnote{Available at {\scriptsize \url{https://riverml.xyz/0.11.1/api/tree/HoeffdingTreeClassifier}}, May 2024.},
 \textsc{hatc}\footnote{Available at {\scriptsize\url{https://riverml.xyz/0.11.1/api/tree/HoeffdingAdaptiveTreeClassifier}}, May 2024.}, and
 \textsc{arfc}\footnote{Available at {\scriptsize \url{https://riverml.xyz/0.11.1/api/ensemble/AdaptiveRandomForestClassifier}}, May 2024.} from River 0.11.1\footnote{Due to computational and time constraints, results were obtained with a balanced subset composed of \num{73770} and \num{49754} samples for the Yelp and MediaWiki data sets, respectively.}. 

Figure \ref{fig:listings} details all hyperparameter optimization values. Their ranges and best values were defined experimentally. Identifying the best values relied on an \textit{ad hoc} implementation of \texttt{\small GridSearch} for data streams.

As the solution operates in streaming mode, no retraining is needed. However, the model's performance is expected to be lower during cold start (initial samples) or with tiny data streams. Consequently, this solution is intended for domains continuously producing large volumes of textual data.

Summing up, the results in tables \ref{tab:classification_results}, \ref{tab:classification_results_d2} and \ref{tab:classification_results_drifts} are estimated with an \textit{ad hoc} implementation of the \texttt{\small progressive\_val\_score}\footnote{Available at {\scriptsize \url{https://riverml.xyz/0.11.1/api/evaluate/progressive-val-score}}, May 2024.} from River 0.11.1. Moreover, the validation scheme comprises prediction and training steps as the system operates in streaming. Consequently, the results displayed correspond to the last computation with the last incoming sample, that is, the last chronologically ordered sample.

Table \ref{tab:classification_results} shows the results obtained in the spam versus non-spam review classification in the four scenarios with the Yelp data set.

\begin{table*}[!thbp]
\centering
\footnotesize
\caption{\label{tab:classification_results}Online spam prediction results (best values in bold) for the Yelp data set.}
\begin{tabular}{ccccccS[detect-weight, mode=text, table-format=3.2]}
\toprule
\bf Scenario & \bf Model & \bf Accuracy & \multicolumn{3}{c}{\bf \textit{F}-measure} & {\bf Time (s)}\\
\cmidrule(lr){4-6}
 & & & Macro & Non-spam & Spam\\
\midrule
\multirow{3}{*}{1} & 
\textsc{htc} & 61.22 & 54.48 & 72.00 & 36.96 & 29.20\\
& \textsc{hatc} & 61.42 & 55.07 & 71.96 & 38.18 & 32.26\\
& \textsc{arfc} & 65.96 & 65.96 & 66.13 & 65.78 & 205.45\\

\midrule

\multirow{3}{*}{2} & 
\textsc{htc} & 62.51 & 57.99 & 71.77 & 44.21 & 5.07 \\
& \textsc{hatc} & 62.17 & 57.70 & 71.44 & 43.97 & 6.39\\
& \textsc{arfc} & 60.76 & 60.75 & 60.99 & 60.52 & 19.93\\

\midrule

\multirow{3}{*}{3} & 
\textsc{htc} & 67.88 & 67.06 & 72.26 & 61.87 & 287.50\\
& \textsc{hatc} & 69.57 & 69.55 & 70.26 & 68.84 & 515.32\\
& \textsc{arfc} & 75.82 & 75.55 & 73.00 & 78.10 & 2346.75\\

\midrule

4 & \textsc{arfc} & \bf 78.75 & \bf 78.44 & \bf 75.85 & \bf 81.03 & 9678.32\\
\bottomrule
\end{tabular}
\end{table*}

In scenarios 1 and 2, the values approach the \SI{60}{\percent} threshold for all models. Unfortunately, the spam \textit{F}-measure in scenario 1 does not reach the \SI{40}{\percent} in \textsc{htc} and \textsc{hatc}. Scenarios 1 and 2 display the same accuracy results since they only differ on the number of running threads. Nonetheless, scenario 3 presents a remarkable improvement in the spam \textit{F}-measure (\num{+30.66} percentage points for \textsc{hatc}). Scenario 3, with data drift detection \& adaptation, reaches a spam \textit{F}-measure of \SI{78.10}{\percent} and an average run-time per sample of \SI{32}{\milli\second} with the \textsc{arfc} model, detecting an average of \num{1.75} drifts per thread (\num{35} data drifts in total). This indicates that data drift detection \& adaptation contributes to increasing the spam classification accuracy (\num{+17.58} percentage points in \textit{F}-measure) and that multi-threading with 20 threads can process an average of \SI{31}{sample\per\second}. Finally, scenario 4 exploits \textsc{arfc}, the best-performing model in scenarios 1, 2, and 3, with data drift detection \& adaptation on a single processing thread. It presents top values for all metrics, including an \SI{81.03}{\percent} in spam \textit{F}-measure and an average run-time per sample of \SI{130}{\milli\second}. This last scenario was able to detect \num{14} drifts and process \SI{8}{sample\per\second}. The difference in the number of data drifts detected in scenario 3 (35) and scenario 4 (14) is caused by thread cold start, \textit{i.e.}, each one of the \num{20} threads starts with a void model.

Table \ref{tab:classification_results_d2} shows the evaluation with the MediaWiki data set. The low results of scenario 2, caused by parallelization, improve in scenario 3 thanks to data drift detection \& adaptation. The promising performance of \textsc{arfc} is further enhanced in scenario 4 with a notable increase in the non-spam \textit{F}-measure between scenario 1 and 4 (\num{+11.84} percentage points). All evaluation metrics are around \SI{85}{\percent}. The number of data drifts and sample processing rate are similar to those obtained with the Yelp data set. In scenario 3, the \textsc{arfc} model reports an average run-time per sample of \SI{47}{\milli\second} (\SI{21}{sample\per\second}) and 38 data drifts (\num{3.8} drifts per thread). The \textsc{ml} model in scenario 4  has identified 10 data drifts and processed \SI{116}{\milli\second\per sample}.

\begin{table*}[!htbp]
\centering
\footnotesize
\caption{\label{tab:classification_results_d2}Online spam prediction results (best values in bold) for the MediaWiki data set.}
\begin{tabular}{ccccccS[detect-weight, mode=text, table-format=3.2]}
\toprule
\bf Scenario & \bf Model & \bf Accuracy & \multicolumn{3}{c}{\bf \textit{F}-measure} & {\bf Time (s)}\\
\cmidrule(lr){4-6}
 & & & Macro & Non-spam & Spam\\
\midrule
\multirow{3}{*}{1} & 
\textsc{htc} & 80.78 & 80.05 & 76.23 & 83.87 & 18.64\\
& \textsc{hatc} & 80.75 & 80.02 & 76.20 & 83.84 & 21.28\\
& \textsc{arfc} & 71.15 & 71.11 & 72.18 & 70.03 & 65.21\\

\midrule

\multirow{3}{*}{2} & 
\textsc{htc} & 79.65 & 78.95 & 75.10 & 82.79 & 4.40\\
& \textsc{hatc} & 79.84 & 79.16 & 75.40 & 82.92 & 5.02\\
& \textsc{arfc} & 69.75 & 69.72 & 70.68 & 68.76 & 9.91\\
\midrule

\multirow{3}{*}{3} & 
 \textsc{htc} & 81.78 & 81.46 & 79.03 & 83.89 & 373.73\\
&\textsc{hatc} & 82.23 & 82.00 & 79.97 & 84.03 & 510.45\\
& \textsc{arfc} & 84.03 & 83.80 & 81.84 & 85.75 & 2333.31\\

\midrule

4 & \textsc{arfc} & {\bf 86.13} & {\bf 85.89} & {\bf 84.02} & {\bf 87.75} & 5817.78\\

\bottomrule
\end{tabular}
\end{table*}

The appropriateness of the proposed drift detection algorithm is supported by its comparison with the Early Drift Detection Method (\textsc{eddm})\footnote{Available at {\scriptsize \url{https://riverml.xyz/0.11.1/api/drift/EDDM}}, May 2024.} and ADaptive WINdowing (\textsc{adwin})\footnote{Available at {\scriptsize \url{https://riverml.xyz/0.11.1/api/drift/ADWIN}}, May 2024.} drift detectors. Table \ref{tab:classification_results_drifts} shows the results of the \textsc{arfc} model in scenario 4 with the three drift detectors and the selected experimental data sets. The proposed drift detector attains the best results followed by \textsc{adwin} (\num{23.24} percent points lower in the \textit{F}-measure for the spam class). Moreover, \textsc{eddm} detects many drifts (793 and 161 for the Yelp and MediaWiki data sets, respectively), which increases the number of training sessions, negatively affecting performance. \textsc{adwin} identifies a few drifts in the Yelp data set (\textit{i.e.}, 6) and a higher number in the MediaWiki data set (38).

\begin{table*}[!htbp]
\centering
\footnotesize
\caption{\label{tab:classification_results_drifts}Online spam prediction results in scenario 4 with different drift detectors (best values in bold).
}
\begin{tabular}{ccccccS[detect-weight, mode=text, table-format=3.2]}
\toprule
\bf Data set & \bf Drift detector & \bf Accuracy & \multicolumn{3}{c}{\bf \textit{F}-measure} & {\bf Time (s)}\\
\cmidrule(lr){4-6}
 & & & Macro & Non-spam & Spam\\
\midrule

\multirow{3}{*}{Yelp} & \textsc{eddm} & 54.58 & 54.58 & 54.53 & 54.63 & 373.37\\
& \textsc{adwin} & 60.56 & 60.56 & 60.70 & 60.42 & 363.12\\
& Proposed & \bf 78.75 & \bf 78.44 & \bf 75.85 & \bf 81.03 & 9678.32\\

\midrule

\multirow{3}{*}{MediaWiki} & \textsc{eddm} & 62.70 & 62.70 & 63.22 & 62.17 & 1078.25\\
& \textsc{adwin} & 65.09 & 65.08 & 65.65 & 64.51 & 1178.08\\
& Proposed & {\bf 86.13} & {\bf 85.89} & {\bf 84.02} & {\bf 87.75} & 5817.78\\

\bottomrule
\end{tabular}
\end{table*}

Analyzing these spam detection results against those of related works found in the literature with the Yelp data set is merely indicative, as it compares the performance of incremental online versus offline classification methods. Nevertheless, the current method outperforms the \SI{62.35}{\percent} accuracy reported by \cite{MOHAWESH2021} by \num{16.4} percent points in the Yelp NYC data set with \num{322167} reviews. Furthermore, the values obtained with the \textsc{adwin} concept drift detection technique by \cite{MOHAWESH2021} are aligned with those reported in Table \ref{tab:classification_results_drifts}. This helps to validate the current method, which attains superior performance. Unfortunately, no information is provided for the specific case of the spam class (\textit{i.e.}, micro-averaging evaluation), in which the current incremental method surpasses the \SI{80}{\percent} barrier in \textit{F}-measure. Moreover, \cite{MOHAWESH2021} focused on concept rather than data drift and did not include explainability capabilities, a distinctive feature of the current method.

\subsection{Explainability}
\label{sec:explainability_results}

\begin{figure*}[!htbp]
\centering
\includegraphics[width=0.8\textwidth]{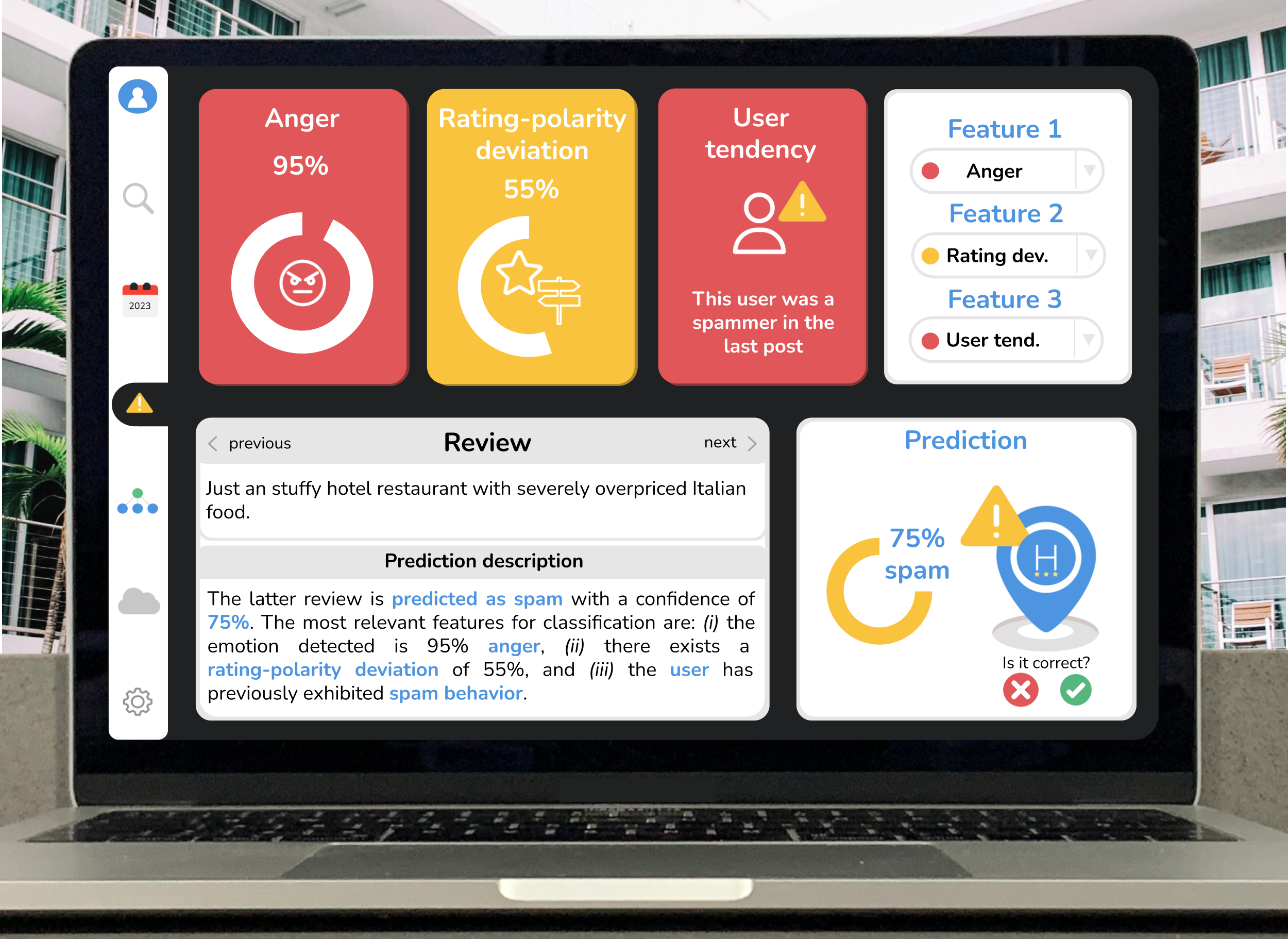}
\caption{\label{fig:dashboard}Explainability dashboard (relevant features).}
\end{figure*}

\begin{figure*}[!htbp]
\centering
\includegraphics[width=0.8\textwidth]{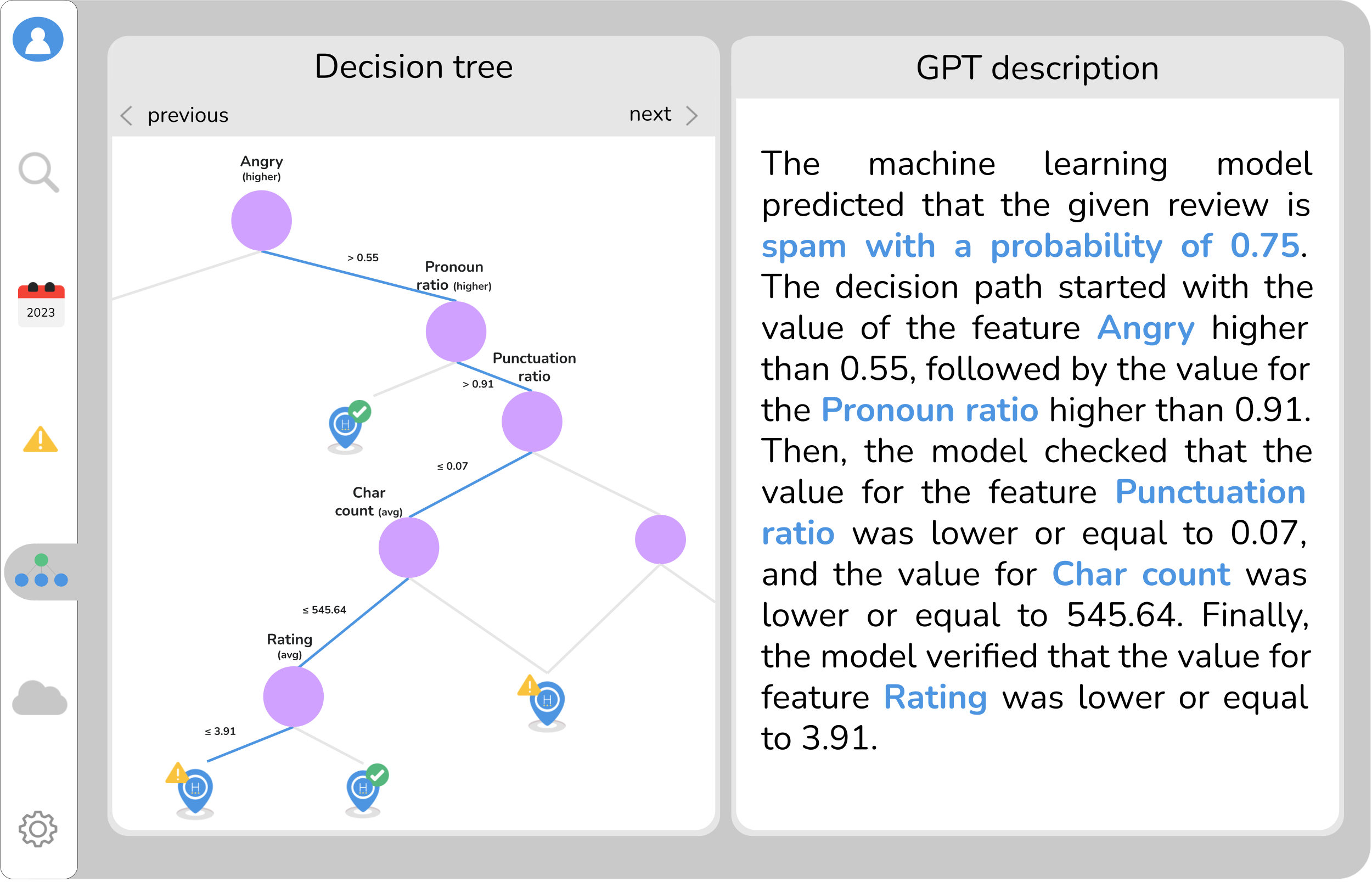}
\caption{\label{fig:dashboard_gtp}Explainability dashboard (decision path and Large Language Model description).}
\end{figure*}

Figure \ref{fig:dashboard} displays the graphical and textual explanation of the classification of an incoming review. The buttons on the left vertical bar enable: (\textit{i}) administrator profile access, (\textit{ii}) search reviews by textual content, (\textit{iii}) search reviews by timestamp, (\textit{iv}) access to alerts, (\textit{v}) visualization of the decision tree and associated natural language description (see Figure \ref{fig:dashboard_gtp}), (\textit{vi}) saving the results in the cloud, and (\textit{vii}) configuring the color layout (\textit{i.e.}, dark or clear mode). The most representative features for the classification are shown in the top part. The relevance of the features corresponds to their frequency of appearance in the decision tree path, considering only positive (greater than) bifurcations (see the graph in Figure \ref{fig:dashboard_gtp}). The white feature navigation panel on the top right displays the most relevant features. The colored circle that accompanies this drop-down menu represents the level of severity (\textit{i.e.}, green when the value is higher than the 50\textsuperscript{th} user quartile, yellow if the feature value is within the 50\textsuperscript{th} - 25\textsuperscript{th} range, and red when it is lower than the 25\textsuperscript{th} user quartile). While these selectors only apply to the colored cards on the left, the review panel on the bottom affects the whole dashboard and enables the analysis of different reviews (\textit{i.e.}, using the previous and next buttons). Finally, there are two additional buttons for feedback (\textit{i.e.}, to indicate whether the prediction is correct or not). This allows a manager to provide feedback, acting as an expert in the loop. The displayed review exhibits a high charge of anger and a significant deviation between the user rating and the detected polarity, the editor has been associated with spam content in the past, and the sample has been classified as spam with a \SI{75}{\percent} confidence using the \texttt{\small Predict\_Proba\_One} function\footnote{Available at {\scriptsize \url{https://riverml.xyz/0.11.1/api/base/Classifier}}, May 2024.} from River 0.11.1. 

Finally, the system presents the decision tree path of the prediction and the corresponding natural language description obtained with \textsc{gpt}3\footnote{Available at {\scriptsize \url{https://openai.com/product}}, May 2024.} (see Figure \ref{fig:dashboard_gtp}). \textsc{gpt}3 was configured to use the \texttt{\small text-davinci-003} model with the default parameters, except the \texttt{\small temperature} parameter, which was set to \num{0.7}, to generate human-like natural language descriptions. At the top, the administrator can navigate the different decision trees using the previous and next buttons, with the decision path highlighted in blue.
 
\section{Conclusion}
\label{sec:conclusion}

The use of crowdsourcing platforms to get information about products and services is growing. Customers search for reviews to make the best decision. Individuals submit dishonest and misleading feedback to manipulate a product or service's reputation or perception. These spam reviews can be created for various reasons, including financial gain, personal grudges, or competitive advantage. To address this problem, the proposed online method identifies and explains spam reviews. In addition, this research contributes with an online explainable classification engine to recognize spam reviews and, thus, to promote trust in digital media.

Specifically, the proposed method comprises (\textit{i}) stream-based data processing (through feature engineering, incremental profiling, and selection), (\textit{ii}) data drift detection \& adaptation, (\textit{iii}) stream-based classification, and (\textit{iv}) explainability. The solution relies on stream-based processing, incrementally updating the profiling and classification models on each incoming event. Specifically, user profiles are computed using user- and content-based features engineered through \textsc{nlp}. Monitoring the incoming streams, the method detects data drifts using static and sliding windows. The classification relies on tree-based classifiers to obtain an interpretable stream-based classification. As a result, the user dashboard includes visual data and natural language knowledge to explain the classification of each incoming event. The experimental classification results of the proposed explainable and stream-based spam detection method show promising performance: \SI{78.75}{\percent} accuracy and \SI{78.44}{\percent} macro \textit{F}-measure obtained with the Yelp data set, and \SI{86.13}{\percent} accuracy and \SI{85.89}{\percent} macro \textit{F}-measure with the MediaWiki data set. Moreover, the proposed data drift detection \& adaptation approach performs better than well-known drift detectors (\num{23.24} percent points higher in the \textit{F}-measure for spam detection). According to the related work analysis, this proposal is the first to jointly provide stream-based data processing, profiling, classification with data drift detection \& adaptation, and explainability.

This solution can be extended to detect orchestrated groups of active spammers thanks to its modular design with \textsc{nlp} techniques and \textit{ad doc} clustering methods for streaming operation. To this end, additional side and content features can be incorporated to cluster contributors by location and temporal affinity. New content-based features can be explored to represent the semantic (\textit{e.g.}, ontology-based like WordNet Domains) and non-semantic similarity (\textit{e.g.}, cosine distance) between reviews. In this regard, the current version of the system already considers sentiment and emotion analysis. The corresponding incremental features can then be designed per user and group of closely related users. The system should, therefore, be able to dynamically adapt to changes in the spamming behavior of both individuals and groups. Moreover, in future work, the online processing throughput can be further improved by adopting parallelization algorithms, which explore the intrinsic distribution of the data together with elastic hardware solutions. Considering the online processing of reviews, the number of threads and the allocation of incoming samples to threads can be location-based, \textit{e.g.}, employing separate dedicated threads to process the reviews of New York, London, or Paris.

\begin{funding}
This work was partially supported by: (\textit{i}) Xunta de Galicia grants ED481B-2021-118 and ED481B-2022-093, Spain; and (\textit{ii}) Portuguese national funds through FCT -- Fundação para a Ciência e a Tecnologia (Portuguese Foundation for Science and Technology) -- as part of project UIDP/50014/2020 (DOI: 10.54499/UIDP/50014/2020 | \small{\url{https://doi.org/10.54499/UIDP/50014/2020}}).
\end{funding}

\section*{Authors' contributions}

\textbf{Francisco de Arriba-Pérez}: Conceptualization, Methodology, Software, Validation, Formnal analysis, Investigation, Resources, Data Curation, Writing - Original Draft, Writing - Review \& Editing, Visualization, Project administration, Funding acquisition. \textbf{Silvia García-Méndez}:  Conceptualization, Methodology, Software, Validation, Formnal analysis, Investigation, Resources, Data Curation, Writing - Original Draft, Writing - Review \& Editing, Visualization, Project administration, Funding acquisition. \textbf{Fátima Leal}: Conceptualization, Resources, Writing - Original Draft. \textbf{Benedita Malheiro}: Conceptualization, Methodology, Validation, Writing - Review \& Editing, Supervision. \textbf{Juan Carlos Burguillo-Rial}: Conceptualization, Writing - Review \& Editing.

\bibliographystyle{infor}
\bibliography{mybibfile}

\begin{biography}\label{bio1}
\author{F. de Arriba-Pérez} received a B.S. degree in telecommunication technologies engineering in 2013, an M.S. degree in telecommunication engineering in 2014, and a Ph.D. in 2019 from the University of Vigo, Spain. He is currently a researcher in the Information Technologies Group at the University of Vigo, Spain. His research includes the development of Machine Learning solutions for different domains like finance and health.
\end{biography}

\begin{biography}\label{bio2}
\author{S. García-Méndez} received a Ph.D. in Information and Communication Technologies from the University of Vigo in 2021. Since 2015, she has worked as a researcher with the Information Technologies Group at the University of Vigo. She is collaborating with foreign research centers as part of her postdoctoral stage. Her research interests include Natural Language Processing techniques and Machine Learning algorithms.
\end{biography}

\begin{biography}\label{bio3}
\author{F. Leal} holds a Ph.D. in Information
and Communication Technologies from the University of Vigo, Spain. She is an Auxiliary Professor at Universidade Portucalense in Porto, Portugal, and a researcher at REMIT (Research on Economics, Management, and Information Technologies). Her research is based on crowdsourced information, including trust and reputation, Big Data, Data Streams, and Recommendation Systems. Recently, she has been exploring blockchain technologies for responsible data processing.
\end{biography}

\begin{biography}\label{bio4}
\author{B. Malheiro} is a Coordinator Professor
at Instituto Superior de Engenharia do Porto, the School of Engineering of the Polytechnic of Porto, and senior researcher at \textsc{inesc} \textsc{tec}, Porto, Portugal. She holds a Ph.D. and an M.Sc. in Electrical Engineering and Computers and a five-year graduation in Electrical Engineering from the University of Porto. Her research interests include Artificial Intelligence, Computer Science, and Engineering Education. She is a member of the Association for the Advancement of Artificial Intelligence (\textsc{aaai}), the Portuguese Association for Artificial Intelligence (\textsc{appia}), the Association for Computing Machinery (\textsc{acm}), and the Professional Association of Portuguese Engineers (\textsc{oe}).
\end{biography}

\begin{biography}\label{bio5}
\author{J. C. Burguillo} Juan C. Burguillo received an M.Sc. degree in Telecommunication Engineering and a Ph.D. degree in Telematics at the University of Vigo, Spain. He is currently a Full Professor at the Department of Telematic Engineering and a researcher at the AtlanTTic Research Center in Telecom Technologies at the University of Vigo. He is the area editor of the journal Simulation Modelling Practice and Theory (SIMPAT), and his topics of interest are intelligent systems, evolutionary game theory, self-organization, and complex adaptive systems.
\end{biography}

\end{document}